\documentclass[letterpaper]{article} 
\usepackage[preprint]{aaai2027}  
\usepackage[hyphens]{url}  
\usepackage{graphicx} 
\usepackage{natbib}  
\usepackage{caption} 
\usepackage{algorithm}
\usepackage{algorithmic}

\usepackage{newfloat}
\usepackage{listings}
\DeclareCaptionStyle{ruled}{labelfont=normalfont,labelsep=colon,strut=off} 
\floatstyle{ruled}
\newfloat{listing}{tb}{lst}{}
\floatname{listing}{Listing}

\usepackage{booktabs}

\newcommand{\sys}{ColSNAP }
\title{\sys: Spatial Matryoshka Training for Multi-Granularity Visual Document Retrieval}
\author {
    Trishan Singha Roy\textsuperscript{\rm 1}\thanks{Work done as an intern at IBM},
    Arkadeep Acharya\textsuperscript{\rm 2},
    Vishwajeet Kumar\textsuperscript{\rm 2},
    Jaydeep Sen\textsuperscript{\rm 2},
    Sachindra Joshi\textsuperscript{\rm 2}
}
\affiliations {
    \textsuperscript{\rm 1}IIT Delhi\\
    \textsuperscript{\rm 2}IBM\\
    ee1230783@iitd.ac.in, acharyarka17@ibm.com, vishk024@in.ibm.com, jaydesen@in.ibm.com, jsachind@in.ibm.com
}

\usepackage[hyphens]{url}  
\usepackage{graphicx} 
\usepackage{natbib}  
\usepackage{caption} 
\usepackage{algorithm}
\usepackage{algorithmic}
\usepackage{todonotes}
\usepackage{amsmath}
\usepackage{amsfonts}
\usepackage{booktabs}
\usepackage{multirow} 

\begin{document}

\maketitle

\begin{abstract}
Multi-modal late-interaction retrievers achieve strong retrieval on visually rich documents by representing each page as per-patch embeddings and matching at the token level. However, this approach incurs high storage costs. Existing compression methods typically fix a single compression level at indexing time, limiting flexibility. We present \sys (Spatial Nested Average Pooling)\footnote{Code will be released upon publication.}, a training method that generates a nested hierarchy of compression levels directly from a backbone's patch grid. By spatially pooling patch embeddings into progressively coarser tiers and training all tiers simultaneously, a single model learns to support retrieval at multiple compression levels without architectural changes. Crucially, a single encoding pass yields every tier, enabling the accuracy-storage trade-off to be configured at indexing time to match available storage budgets, rather than being fixed during training. We demonstrate that models trained using \sys maintain near full-resolution retrieval performance under substantial compression and that \sys transfers effectively across multiple late-interaction backbones, and achieves most of its improvements via a lightweight adaptation stage applied to a pre-trained retriever.

\end{abstract}

\section{Introduction}

Visual Document Retrieval (VDR) retrieves document pages relevant to a natural
language query using both the visual layout and textual content of a page
\citep{yan2026unlockingmultimodaldocumentintelligence,
faysse2025colpaliefficientdocumentretrieval}. Relevance cannot be determined from
text alone: it depends on visual elements such as tables, charts, and figures, and on
their spatial organization. Early approaches applied optical character recognition
(OCR)~\citep{hegghammer_ocr_2022, zhang2025ocrhindersragevaluating} to extract text
for a conventional retriever, but such pipelines are brittle, as recognition errors
propagate downstream and layout structure is lost when pages are verbalized.
Vision-language models (VLMs) instead embed the rendered page image directly,
eliminating the extraction stage.

The emergence of multi-modal embedding models has led to two dominant approaches.
1) \textbf{Single-vector methods} encode each page and query into one dense embedding,
scored by cosine similarity \citep{ma2024unifyingmultimodalretrievaldocument,
meng2025vlm2vecv2advancingmultimodalembedding}. These are compact and support efficient
approximate nearest-neighbour search, but compressing a page into one vector discards
the fine-grained evidence that complex documents require, where relevance may hinge on
a single table row or figure annotation
\citep{weller2026theoreticallimitationsembeddingbasedretrieval}.
2) \textbf{Multi-vector methods} adopt the late-interaction paradigm of ColBERT
\citep{khattab2020colbertefficienteffectivepassage}, extending it to visual documents.
Systems such as ColQwen and ColPali~\citep{faysse2025colpaliefficientdocumentretrieval}
represent a page as per-patch embeddings and score queries with a MaxSim operator that
aligns each query token to its most relevant page region, preserving the localized
evidence that pooling discards. This has established multi-vector representations as
the dominant paradigm in VDR \citep{moreira2026nemotroncolembedv2topperforming,
gunther2025jinaembeddingsv4universalembeddingsmultimodal}. The cost, however, is
substantial: a single page yields hundreds to thousands of stored vectors, so index
size and query-time scoring both scale with the number of patches per page, and
higher-resolution inputs push this further, making storage a limiting factor in
large-scale deployment. This cost--quality tradeoff motivates two research questions:
\begin{itemize}
\item \textbf{RQ1:} Can a method maintain the retrieval performance of an existing retriever while substantially reducing the number of vectors needed to encode each document page?
\item \textbf{RQ2:} Does such a method generalize across retrievers of differing size and retrieval performance?
\end{itemize}
We address these questions with \textbf{\sys} (Spatial Nested Average Pooling), a training recipe that builds a nested hierarchy of compression levels directly from the backbone's patch grid. The key insight is that pretrained patch grids exhibit natural hierarchy-spatially adjacent patches are highly redundant, so averaging them yields meaningful coarser embeddings. In a single encoding pass, we spatially pool patch embeddings into a pyramid of granularities ($1024 \to 128 \to 16 \to 1$), adding no tokens and no change to the encoder architecture. All tiers are trained jointly via Spatial Matryoshka learning with two distillation signals: (1) a frozen-teacher term anchoring all tiers to the original pretrained model, and (2) a cross-level term aligning coarse tiers to the live model's finest tier. At indexing time, users select a tier matching their storage budget, achieving up to $\sim\!70\times$ smaller storage footprint on Nemotron ColEmbed 4B~\citep{moreira2026nemotroncolembedv2topperforming} and $\sim\!130\times$ smaller on ColPali v1.3~\citep{faysse2025colpaliefficientdocumentretrieval}.

\begin{figure*}[t]
  \centering
  \includegraphics[width=\textwidth]{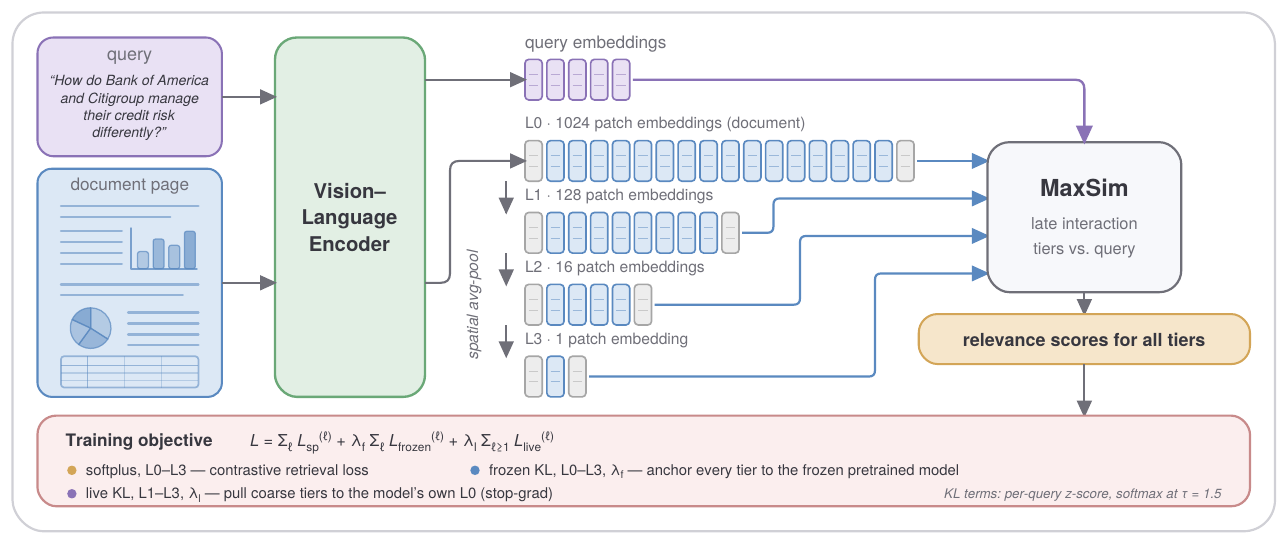}
  \caption{Overview of ColSNAP.
  A shared vision--language encoder maps a document page to a row of patch
  embeddings (blue), flanked by a few auxiliary template tokens (grey), and the
  query to a set of query embeddings (purple). Spatial average-pooling compresses
  the patch embeddings into a nested pyramid of tiers (L0--L3) while leaving the
  auxiliary tokens unchanged; any tier can then be scored against the query with
  late-interaction MaxSim. A single encoding pass produces every tier, so the
  compression level is selected at index time.}
  \label{fig:pipeline}
\end{figure*}

We summarize our contributions as follows:
\begin{itemize}
\item We propose \sys{}, a method that reduces storage costs of multi-vector retrievers by learning multiple compression levels from a single patch grid, without changing the encoder architecture.

\item \sys{} achieves up to $\sim\!70\times$ compression while retaining over $90\%$ of full-resolution quality, with compression levels configurable at indexing time.

\item The method generalizes across retrievers of differing size and retrieval performance (demonstrated on Nemotron ColEmbed 4B~\citep{moreira2026nemotroncolembedv2topperforming} and ColPaLI v1.3~\citep{faysse2025colpaliefficientdocumentretrieval}) and requires only lightweight adaptation of pre-trained models.

\item We evaluate on ViDoRe v1~\citep{faysse2025colpaliefficientdocumentretrieval}, v2~\citep{mace2025vidorebenchmarkv2raising}, and the public splits of v3~\citep{loison2026vidorev3comprehensiveevaluation} benchmarks and show that multi-vector retrieval is practical at scale with \sys{}.
\end{itemize}

The remainder of this paper is organized as follows. Section~\ref{sec:related_work} reviews related work in neural retrieval and representation learning. Section~\ref{sec:method} presents the \sys method in detail. Section~\ref{sec:experimental_setup} describes our experimental setup and benchmarks. Section~\ref{sec:results} presents results and analysis. Finally, Section~\ref{sec:conclusion} concludes the paper.
\section{Related Work}
\label{sec:related_work}
Late-interaction retrieval, introduced by ColBERT
\citep{khattab2020colbertefficienteffectivepassage}, encodes queries and documents
into sequences of per-token embeddings and scores a pair with MaxSim, preserving
fine-grained matching that a single pooled vector discards, at the cost of one vector
per token. ColPali \citep{faysse2025colpaliefficientdocumentretrieval} extended this
to visually rich documents by encoding rendered page images into multi-vector patch
embeddings, sidestepping brittle OCR pipelines, and the same recipe has since carried
over to stronger backbones such as Nemotron ColEmbed
\citep{moreira2026nemotroncolembedv2topperforming}. However, this accuracy comes at a
cost: each page is stored as hundreds to thousands of patch vectors, so index size
and retrieval latency grow with the number of tokens per page.

A large body of work reduces this cost for text. ColBERTv2
\citep{santhanam2022colbertv2effectiveefficientretrieval} quantizes embeddings
against learned centroids, later pruned by PLAID
\citep{santhanam2022plaidefficientenginelate}; other methods reduce the number of
stored vectors directly, via token pooling
\citep{clavie2024reducingfootprintmultivectorretrieval}, learned token retention
\citep{lee2024rethinkingroletokenretrieval}, clustering
\citep{veneroso2025crispclusteringmultivectorrepresentations}, or a single
fixed-dimensional encoding whose inner product approximates MaxSim
\citep{dhulipala2026muveramultivectorretrievalfixed}. Several of the strongest
methods, such as CITADEL's lexical-key routing \citep{li-etal-2023-citadel}, exploit
discrete, term-level structure that has no analogue for continuous image patches.

Compression for visual document retrieval is more recent, from training-free pruning
and merging of patch embeddings
\citep{yan2025docprunerstorageefficientframeworkmultivector,
liu2026structuralanchorpruningtrainingfree, yan2026sculptingvectorspaceefficient,
yan2026visuallatechunkingempirical} to methods that finetune a module or the
retriever itself \citep{ma2025storageefficientvisualdocumentretrieval,
cha2026reinpoolreinforcementlearningpooling, park2026visualtokensmatterequally}. Most
of these commit to a single compression level fixed at indexing time. A few instead
serve several levels from one model: MM-Matryoshka nests along embedding width and
encoder depth \citep{xiang2026mmmatryoshkabudgetelasticvisualdocument}, MetaEmbed
appends learnable Meta Tokens trained end to end in nested coarse-to-fine groups
\citep{xiao2026metaembedscalingmultimodalretrieval}, and jina-embeddings-v4 trains
adapters to emit both a truncatable single vector and a multi-vector representation
\citep{gunther2025jinaembeddingsv4universalembeddingsmultimodal}. The spatial layout
of the patch grid, the one form of structure visual tokens reliably possess, remains
largely unused as an axis for such a hierarchy.

Matryoshka Representation Learning (MRL)
\citep{kusupati2024matryoshkarepresentationlearning} encodes multiple granularities
within a single vector by ordering dimensions by importance. This principle has been extended from dimension to vector count:M3
\citep{cai2024matryoshkamultimodalmodels} learns visual tokens for multimodal generation, while MetaEmbed
\citep{xiao2026metaembedscalingmultimodalretrieval} applies it to late-interaction retrieval via Meta Tokens. To the best of our knowledge, for visual document retrieval, ColSNAP is the first method to build a nested multi-vector hierarchy from the backbone's patch grid. We spatially average-pool per-patch embeddings into a granularity pyramid, training each tier for retrieval. Since this spatial axis is orthogonal to embedding dimensionality, it composes with dimension-side compression without changing the encoder.
\section{\sys}
\label{sec:method}

\subsection{Overview}
\sys addresses the storage challenge of multi-vector retrievers by constructing a
nested hierarchy of token compression levels from a single forward pass. The core insight
is simple: a pretrained $32 \times 32$ patch grid exhibits natural spatial hierarchy.
Adjacent patches encode redundant information about the same document region, so
spatially averaging them substantially reduces the vector count while retaining much
of the retrieval quality. \sys trains all compression tiers jointly using a shared contrastive objective, together with two distillation signals. This lets the accuracy-storage trade-off be configured at indexing time rather than fixed during training. This allows a single model to serve diverse deployment scenarios without
retraining.

\subsection{Setting and Background}

\paragraph{Problem:}
Given a textual query and a corpus of documents whose pages are represented as
rendered images, Visual Document Retrieval ranks pages by their relevance to the
query without OCR or layout parsing.

\paragraph{Encoding:}
A vision-language model encodes both sides into a shared $d$-dimensional space. The
query is tokenized and encoded into a set of query embeddings
$Q = \{q_1, \dots, q_{n_q}\}$, $q_t \in \mathbb{R}^d$. Each page image is processed
by a vision encoder into a set of document embeddings. The encoder emits a
$32\times32$ grid of patch embeddings, denoted by $p_1, \dots, p_{n_p}$ where
$n_p = 1024$ and $p_i \in \mathbb{R}^d$, as well as a small number of auxiliary
tokens (e.g., special tokens before and after the patch grid), denoted
$a_1, \dots, a_k$ where $a_j \in \mathbb{R}^d$. The full document representation is:
\begin{equation}
D = \{a_1, \dots, a_k,\; p_1, \dots, p_{n_p}\}.
\end{equation}

\paragraph{Late Interaction Scoring:}
Late-interaction retrieval \citep{khattab2020colbertefficienteffectivepassage} scores
a query-document pair by summing, over each query embedding, its maximum similarity
to any document embedding:
\begin{equation}
S(Q, D) \;=\; \sum_{t=1}^{n_q} \; \max_{v \in D} \; \langle q_t, v \rangle ,
\label{eq:tiered-maxsim}   
\end{equation}
where $|D| = k + n_p$. This preserves fine-grained detail by matching each query
token to its most relevant document region.

\subsection{Spatial Pooling Pyramid}
\begin{figure}[t]
  \centering
  \includegraphics[width=1\linewidth]{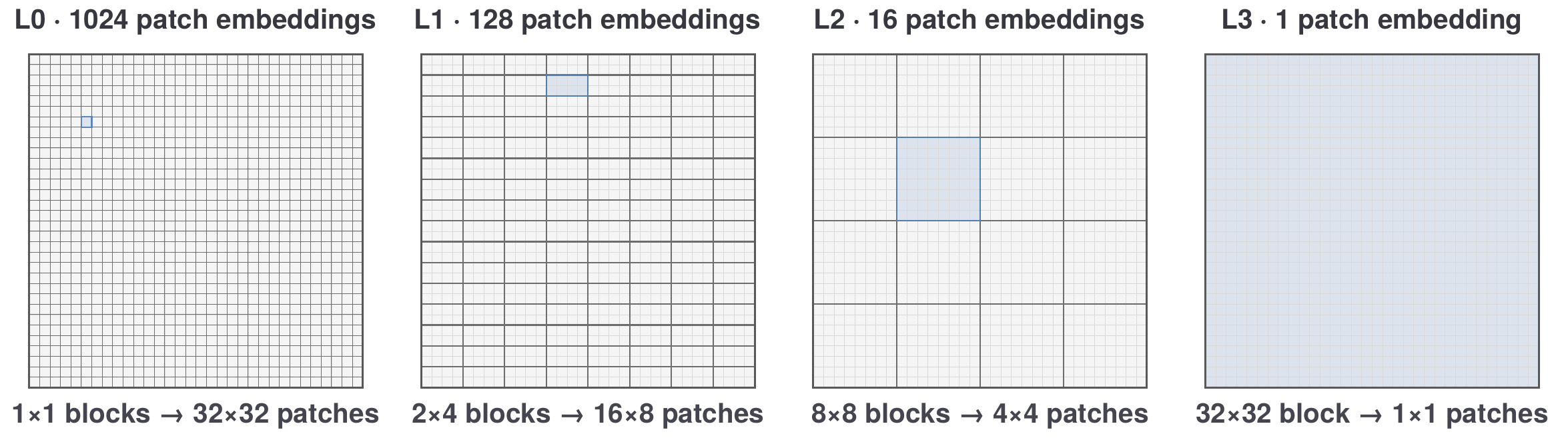}
  \caption{Spatial pooling pyramid.
  The $32\times32$ patch grid is average-pooled over increasingly large windows to
  form four tiers: L0 ($1\times1$ blocks, $1024$ embeddings), L1 ($2\times4$ blocks,
  $16\times8=128$), L2 ($8\times8$ blocks, $4\times4=16$), and L3 (a single
  $32\times32$ block, $1$). Each shaded block is averaged into one pooled embedding.}
  \label{fig:pyramid}
\end{figure}
\sys builds a \emph{nested} pyramid of coarser views of the $32 \times 32$ patch grid
by spatially average-pooling patch embeddings over progressively larger windows.
Because each pooling window is nested inside the next, the resulting tiers are
hierarchically consistent: every coarse embedding summarizes a contiguous spatial
block of the finer embeddings beneath it.

This construction produces four tiers, denoted $L_0, L_1, L_2, L_3$ (indexed by
$\ell \in \{0, 1, 2, 3\}$), where tier $L_\ell$ contains $n_\ell$ patch embeddings
arranged as a grid:
\begin{equation}
\begin{aligned}
L_0 &: 32 \times 32 = 1024, &\quad L_1 &: 16 \times 8 = 128, \\
L_2 &: 4 \times 4 = 16,     &\quad L_3 &: 1 \times 1 = 1 .
\end{aligned}
\end{equation}
Each tier is a complete, coarser summary of the same underlying patches. We denote
the pooled patch embeddings at tier $\ell$ as
$p_1^{(\ell)}, \dots, p_{n_\ell}^{(\ell)}$. Importantly, the coarser tiers introduce
no additional parameters and reuse the same underlying embeddings;
Figure~\ref{fig:pyramid} illustrates the construction.


\subsection{Multi-Tier Scoring}
\sys scores identically across all tiers using the MaxSim operator. At tier $L_\ell$,
we replace the $n_p = 1024$ patch embeddings with their pooled counterparts
$p_1^{(\ell)}, \dots, p_{n_\ell}^{(\ell)}$, while the auxiliary tokens
$a_1, \dots, a_k$ are carried over unchanged. The document representation at tier
$\ell$ is:
\begin{equation}
D^{(\ell)} = \{a_1, \dots, a_k,\; p_1^{(\ell)}, \dots, p_{n_\ell}^{(\ell)}\}.
\end{equation}
The relevance score at tier $\ell$, written $S^{(\ell)}(Q, D)$, is then
Eq.~\ref{eq:tiered-maxsim} with $D$ replaced by $D^{(\ell)}$, where
$|D^{(\ell)}| = k + n_\ell$. $\ell = 0$ applies no pooling, so $D^{(0)} = D$: the finest tier is identical to the original, unpooled retriever. We compress only the patch embeddings, whose
count varies across tiers through $1024, 128, 16, 1$, leaving the auxiliary tokens
untouched.

\subsection{Joint Training via Spatial Matryoshka Learning}
To optimize all tiers simultaneously without degrading the original model's
performance, we employ a joint training objective combining three loss terms.

\paragraph{Spatial Matryoshka Ranking Loss.}
Each tier is trained to rank matched documents above in-batch negatives. During
training, each step samples a batch of $B$ matched query--document pairs. For every
tier $L_\ell$ (where $\ell \in \{0, 1, 2, 3\}$), we compute MaxSim scores between all
$B$ queries and $B$ documents, forming a score matrix:
\begin{equation}
S^{(\ell)} \in \mathbb{R}^{B \times B},\quad
S^{(\ell)}_{ij} = S^{(\ell)}(Q_i, D_j^{(\ell)}),
\end{equation}
where each entry is the MaxSim score of Eq.~\ref{eq:tiered-maxsim}, evaluated for
query $i$ against document $j$ at tier $\ell$.
The diagonal entries $S^{(\ell)}_{ii}$ contain scores for matched (positive) pairs,
and off-diagonal entries $S^{(\ell)}_{ij}$ (where $i \neq j$) are in-batch negatives.
For query $i$ at tier $\ell$, the positive score is the diagonal entry
$s_+^{(i,\ell)} = S^{(\ell)}_{ii}$, and the hardest negative is
$s_-^{(i,\ell)} = \max_{j \neq i} S^{(\ell)}_{ij}$.
A softplus penalty pushes the positive above this hardest negative, and the Spatial
Matryoshka ranking loss sums this across all tiers:
\begin{equation}
\mathcal{L}_{\text{SMRL}} = \sum_{\ell=0}^{3} \frac{1}{B}\sum_{i=1}^{B}
\log\!\bigl(1 + \exp\!\bigl(s_-^{(i,\ell)} - s_+^{(i,\ell)}\bigr)\bigr).
\end{equation}

In order to preserve the performance of the model at its original resolution, we add
a frozen-teacher term $\mathcal{L}^{(\ell)}_{\text{frozen}}$, which matches the
student's distribution over the $B$ in-batch documents to that of a frozen copy of
the pretrained backbone through a temperature-scaled KL divergence. It is applied at
every level, anchoring each tier to the original model at full resolution.

Additionally, in order to align the coarser tiers with the model's own finest tier,
we introduce a live cross-tier term $\mathcal{L}^{(\ell)}_{\text{live}}$, which
matches the student's distribution at tier $\ell$ to its current L0 distribution,
detached from the gradient, and applies at $\ell \in \{1,2,3\}$. In both terms the
per-query scores are z-scored before the softmax, so that differences in score scale
across tiers do not dominate the distributions being matched.

Finally, we obtain the overall objective as
\begin{equation}
\mathcal{L} = \mathcal{L}_{\text{SMRL}}
\;+\; \lambda_{f}\sum_{\ell=0}^{3}\mathcal{L}^{(\ell)}_{\text{frozen}}
\;+\; \lambda_{l}\sum_{\ell=1}^{3}\mathcal{L}^{(\ell)}_{\text{live}} .
\label{eq:objective}
\end{equation}

\subsection{Flexible Deployment via Granularity Selection}
The nested design decouples the compression level from training. Because all tiers
are optimized jointly and remain hierarchically consistent, a corpus encoded at $L_0$
can be pooled down to any coarser tier on demand. At indexing time, each page is
stored at the tier $L_\ell$ that fits the available storage budget. At query time,
retrieval cost is bounded by that choice: MaxSim operates over $k + n_\ell$
embeddings, where the auxiliary count $k$ is fixed and only $n_\ell$ varies. A single
trained model therefore serves multiple operating points without retraining or
maintaining separate checkpoints.
\section{Experimental Setup}
In this section we describe the backbone models adapted with \sys, the datasets and training configurations used in our experiments, and the benchmarks and evaluation metrics used for evaluation.
\label{sec:experimental_setup}
\subsection{Model Backbones}

We apply \sys to two late-interaction retrieval backbones with different scales and pretraining. Our primary backbone is \textbf{Nemotron ColEmbed VL 4B} \citep{moreira2026nemotroncolembedv2topperforming}, which produces 2560-dimensional embeddings and ranks among the top models on the ViDoRe V3 leaderboard. We also apply \sys to \textbf{ColPali v1.3} \citep{faysse2025colpaliefficientdocumentretrieval}, built on PaliGemma-3B \citep{beyer2024paligemmaversatile3bvlm}, which produces 128-dimensional embeddings. As a smaller, widely used backbone, ColPali helps to assess the generalizability of \sys across model scales.

For both backbones, input images are resized to produce a $32\times32$ grid of patch tokens, enabling the same pooling pyramid to be used without modification. In addition to the patch tokens, Nemotron and ColPali produce $14$ and $7$ auxiliary tokens respectively, which are preserved unchanged across all compression tiers.

\subsection{Training Setup}

\paragraph{Training data:}
We use the ColPali-Train dataset \citep{faysse2025colpaliefficientdocumentretrieval} with 95:5 train:validation split. In our primary experiments, each training step samples 128 matched query–document pairs (positives). To explore the effect of hard negatives, we additionally conduct experiments using mined hard negatives released by Nomic AI \citep{nomicai2025colpaliqueriesminedbysource}\footnote{\url{https://huggingface.co/datasets/nomic-ai/colpali-queries-mined-20250321-by-source}}, where each query is paired with one hard negative sampled uniformly from its top-10 mined candidates. In this setting, the effective batch size is 128 documents (64 positives + 64 negatives).  
\paragraph{Optimization and hyperparameters:}
We jointly optimize all the four tiers using the objective in Eq.~\ref{eq:objective} for 5 epochs, selecting the best checkpoint on the validation split. Training uses LoRA \citep{hu2021loralowrankadaptationlarge} ($\mathrm{rank}=32$, $\alpha=32$) with AdamW \citep{loshchilov2019decoupledweightdecayregularization} and a linear warmup-decay schedule. Models are trained in bfloat16 with distributed data parallelism on four NVIDIA A100 GPUs using an effective batch size of 128. Unless otherwise stated, we set $\lambda_f=2$ for Nemotron and $\lambda_f=0.5$ for ColPali, with $\lambda_l=0.3$ and a distillation temperature of $\tau=1.5$ for both backbones.

\subsection{Evaluation Benchmarks and Metrics}

\paragraph{Benchmarks.}
We evaluate the performance of models adapted using \sys on ViDoRe v1 \citep{faysse2025colpaliefficientdocumentretrieval}, ViDoRe v2 \citep{mace2025vidorebenchmarkv2raising}, and the public splits of ViDoRe V3 \citep{loison2026vidorev3comprehensiveevaluation}. ViDoRe V1 is an in-distribution benchmark, while ViDoRe v2 and v3 evaluate out-of-distribution across document types and domains.

\paragraph{Metrics.}
Following the official leaderboards, we report nDCG@5 on ViDoRe v1/v2 and nDCG@10 on ViDoRe v3. Each pyramid level (L0--L3) is evaluated independently using tiered MaxSim (Eq.~\ref{eq:tiered-maxsim}) within the MTEB framework \citep{muennighoff2023mtebmassivetextembedding, enevoldsen2025mmtebmassivemultilingualtext}. We also report \emph{retention}, the ratio of a tier's retrieval score to that of the corresponding backbone at full resolution (L0).

\section{Results and Analysis}
\label{sec:results}
Our analysis proceeds by evaluating \sys across two late-interaction backbones with different scales and architectures: (i) Nemotron ColEmbed 4B and (ii) ColPali v1.3. We first discuss retrieval quality and compression behavior across both backbones, and then highlight gains \sys delivers in training efficiency, storage requirements, and retrieval cost.

\subsection{Performance of ColSNAP on Nemotron ColEmbed  backbone}
\label{sec:Nemotro_results}
\begin{figure}[t]
  \centering
  \includegraphics[width=0.96\linewidth]{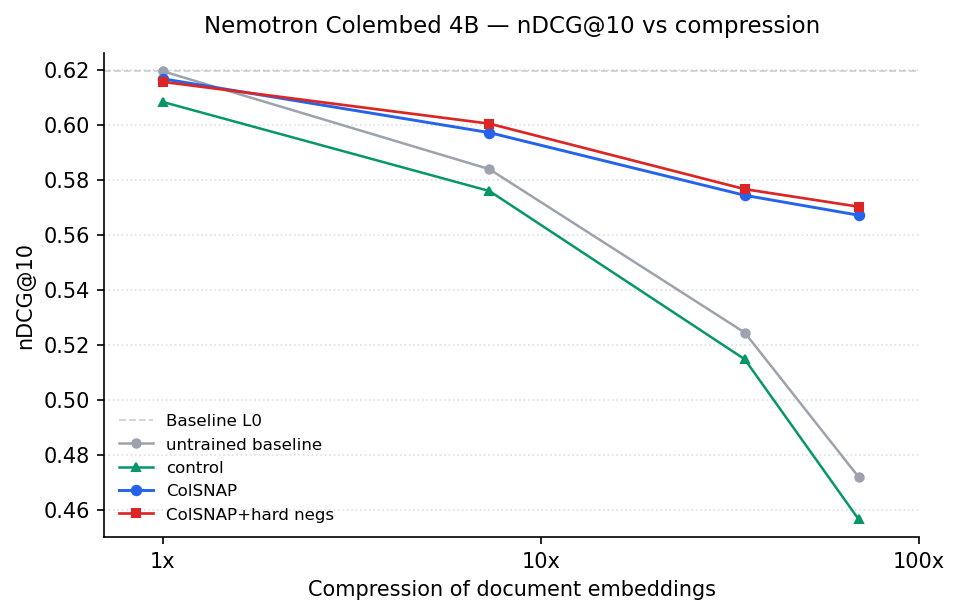}
  \caption{Compression--quality tradeoff on Nemotron (ViDoRe V3, nDCG@10).
  Compression on a log scale; each marker is a tier (L0/L1/L2/L3 at $1\times$,
  $7.3\times$, $34.6\times$, $69.2\times$). Curves show the untrained baseline, a
  full-resolution-only control, ColSNAP, and ColSNAP with mined hard negatives.}
  \label{fig:nemotron_curve}
\end{figure}

We evaluate a model trained with \sys on the Nemotron ColEmbed VL 4B backbone and report results in Table~\ref{tab:leaderboard} alongside existing late-interaction baselines of similar parameter size on ViDoRe v1, v2, and v3. In Figure~\ref{fig:nemotron_curve}, we further illustrate the compression--quality trade-off on this backbone, comparing four settings: (i) the untrained model with our pooling applied at inference time, (ii) the model finetuned with only the softplus loss at full resolution, (iii) the model trained with \sys using our proposed objective in Equation~\ref{eq:objective}, and (iv) the model trained with \sys using Equation~\ref{eq:objective} and data containing mined hard negatives. Based on these results, we highlight three key observations:

\paragraph{1) Retrieval quality degrades gracefully with compression:}
Figure~\ref{fig:nemotron_curve} shows that model trained with \sys achieves a graceful compression--quality trade-off. On ViDoRe v3, retrieval quality decreases from 61.7 at full resolution (L0) to 59.7, 57.4, and 56.7 for the L1, L2, and L3 tiers, respectively. Notably, the L1 tier stores roughly seven times fewer vectors while retaining over 96\% of the full-resolution performance. In contrast, both the untrained baseline and the full-resolution control exhibit a much steeper degradation from L0 to L3, highlighting that the proposed training objective in \sys is key to learning compressed representations that retain retrieval quality across all the tiers.

\paragraph{2) Compressed tiers remain competitive with full-resolution retrievers.}
Table~\ref{tab:leaderboard} shows that L1 tier surpasses several full-resolution retrievers on ViDoRe v3, including jina-embeddings-v4 \citep{gunther2025jinaembeddingsv4universalembeddingsmultimodal} and colnomic-embed-multimodal-3b \citep{nomicembedmultimodal2025} across the ViDoRe benchmarks. This holds even for more compressed tiers: on ViDoRe v3, L2 (57.4) nearly matches jina-embeddings-v4 (57.5) and outperforms colnomic-embed-multimodal-3b (56.4). Even single patch-embedding L3 (56.7) stays competitive on this benchmark, beating colnomic-embed-multimodal-3b and trailing llama-nemoretriever-colembed-3b-v1 (57.1) by just 0.4 nDCG. A similar trend can be observed across ViDoRe V1 and ViDoRe V2, thus underscoring that \sys achieves substantial compression with minimal quality loss versus strong full-resolution baselines.

\paragraph{3) Compression remains robust under distribution shift. }
As shown by the retention values in Table~\ref{tab:leaderboard}, on the in-distribution ViDoRe v1 benchmark, even the single patch-embedding L3 tier retains 97.2\% of the full-resolution performance. This retention remains high on the out-of-distribution ViDoRe v2 and v3 benchmarks, with L3 preserving 92.5\% and 91.5\% of full-resolution performance, respectively. Across all benchmarks, the L1 tier is particularly robust, retaining at least 95.4\% of the full-resolution score. These results indicate that \sys helps in learning compressed representations that generalize well beyond the training data distribution.

Together, these observations address \textbf{RQ1}: a retriever adapted with \sys preserves over $91\%$ of its full-resolution quality while storing up to 69× fewer embeddings per page.

\begin{table*}[t]
\centering
{\small
\setlength{\tabcolsep}{3.5pt}
\begin{tabular}{ll cc cccccccc c}
\toprule
\multicolumn{2}{l}{\multirow{2}{*}{Model}} & \multicolumn{2}{c}{ViDoRe v1/v2 (nDCG@5, \%)} & \multicolumn{9}{c}{ViDoRe v3 public (nDCG@10, \%)} \\
\cmidrule(lr){3-4}\cmidrule(lr){5-13}
\multicolumn{2}{l}{} & v1 & v2 & CS & Ene & FiEn & FiFr & HR & Ind & Pha & Phy & Mean \\
\midrule
\multicolumn{2}{l}{ColQwen3.5-4.5B} & 91.5 & 64.3 & 78.7 & 68.0 & 64.1 & 48.6 & 62.1 & 55.2 & 65.6 & 50.3 & 61.6 \\
\multicolumn{2}{l}{jina-embeddings-v4} & 90.4 & 58.2 & 71.8 & 63.5 & 59.3 & 46.1 & 59.5 & 50.4 & 63.1 & 46.6 & 57.5 \\
\multicolumn{2}{l}{llama-nemoretriever-colembed-3b-v1} & 91.0 & 63.3 & 75.2 & 62.1 & 60.9 & 43.8 & 58.7 & 47.1 & 63.7 & 45.1 & 57.1 \\
\multicolumn{2}{l}{colnomic-embed-multimodal-3b} & 89.9 & 55.7 & 72.7 & 64.5 & 56.3 & 44.3 & 57.3 & 47.4 & 61.1 & 47.6 & 56.4 \\
\midrule
\multicolumn{2}{l}{Nemotron ColEmbed 4B} & 91.7$_{(100.0)}$ & 65.0$_{(100.0)}$ & 78.6 & 68.4 & 65.9 & 49.5 & 62.8 & 54.6 & 66.6 & 49.2 & 62.0$_{(100.0)}$ \\
\midrule
\multirow{4}{1.7cm}{ColSNAP-Nemotron-ColEmbed-4B} & L0 & 91.7$_{(100.0)}$ & 63.2$_{(97.2)}$ & 79.0 & 67.6 & 65.0 & 49.4 & 62.9 & 53.9 & 66.0 & 49.5 & 61.7$_{(99.6)}$ \\
 & L1 & 90.5$_{(98.7)}$ & 62.0$_{(95.4)}$ & 77.8 & 66.2 & 60.9 & 48.0 & 60.2 & 50.8 & 64.6 & 49.4 & 59.7$_{(96.4)}$ \\
 & L2 & 89.7$_{(97.8)}$ & 61.0$_{(93.7)}$ & 76.7 & 64.4 & 58.5 & 45.3 & 56.6 & 47.0 & 63.1 & 48.0 & 57.4$_{(92.7)}$ \\
 & L3 & 89.1$_{(97.2)}$ & 60.2$_{(92.5)}$ & 75.6 & 63.3 & 57.7 & 44.8 & 56.0 & 46.1 & 62.0 & 48.1 & 56.7$_{(91.5)}$ \\
\midrule
\multirow{4}{1.8cm}{ColSNAP-Nemotron-ColEmbed-4B-hard-negs} & L0 & 91.9$_{(100.2)}$ & 63.4$_{(97.5)}$ & 78.5 & 68.2 & 64.0 & 49.3 & 62.8 & 53.8 & 65.7 & 50.2 & 61.6$_{(99.4)}$ \\
 & L1 & 91.2$_{(99.4)}$ & 62.2$_{(95.7)}$ & 77.7 & 66.8 & 61.2 & 47.7 & 60.4 & 51.2 & 65.0 & 50.3 & 60.0$_{(96.9)}$ \\
 & L2 & 90.4$_{(98.6)}$ & 60.5$_{(93.0)}$ & 77.0 & 65.0 & 57.0 & 44.5 & 57.9 & 48.0 & 62.8 & 49.2 & 57.7$_{(93.1)}$ \\
 & L3 & 90.1$_{(98.3)}$ & 58.6$_{(90.1)}$ & 76.1 & 64.2 & 55.9 & 43.6 & 57.6 & 47.1 & 62.3 & 49.4 & 57.0$_{(92.0)}$ \\
\bottomrule
\end{tabular}}
\caption{Comparison across ViDoRe generations. Mean nDCG@5 on ViDoRe v1 and v2, and per-dataset nDCG@10 on the eight public ViDoRe v3 tasks (computer science, energy, finance-en, finance-fr, HR, industrial, pharmaceuticals, physics), with their mean. Parenthesised subscripts denote retention, computed relative to the untrained Nemotron ColEmbed 4B baseline. Top: representative models (retention not applicable). Bottom: the baseline backbone trained with ColSNAP at each tier (L0--L3), without and with mined hard negatives.}
\label{tab:leaderboard}
\end{table*}

\subsection{Transferability of \sys to ColPali v1.3 backbone}
\label{sec:colpali}
\begin{figure}[t]
  \centering
  \includegraphics[width=0.96\linewidth]{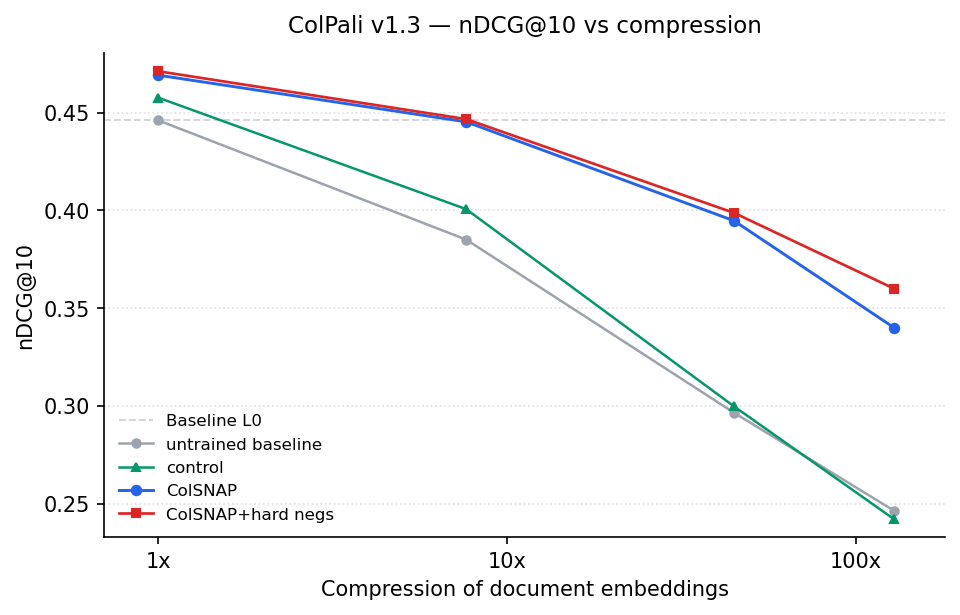}
  \caption{Compression--quality tradeoff on ColPali (ViDoRe V3, nDCG@10).
  Same axes and same four settings as Figure~\ref{fig:nemotron_curve} (tiers at
  $1\times$, $7.6\times$, $44.8\times$, $128.9\times$).}
  \label{fig:colpali_curve}
\end{figure}
We evaluate the model trained with \sys on ColPali v1.3 to assess the transferability of the method to a smaller and structurally different backbone under the same experimental setting as described earlier for Nemotron ColEmbed backbone in subsection \ref{sec:Nemotro_results}. The key metrics have been highlighted in Table~\ref{tab:colpali_vidore} and Figure~\ref{fig:colpali_curve}. We make three key observations:

\paragraph{1) \sys transfers effectively to smaller backbones.}
Similar to Nemotron, ColPali exhibits a graceful compression--quality trade-off under \sys. The L1 tier reduces the number of embedding vectors by $7.6\times$ while remaining nearly lossless, retaining $100.2\%$, $98.0\%$, and $99.8\%$ of the full-resolution quality on ViDoRe v1, v2, and v3, respectively. This demonstrates that \sys generalizes across backbones and preserves retrieval quality under moderate compression. As on Nemotron, the full-resolution control leaves the coarse tiers near their untrained values, indicating that the improvements at the compressed tiers likewise come from the nested objective rather than additional training alone.

\paragraph{2) Extreme compression is limited by backbone capacity.}
At more aggressive compression levels, ColPali shows a larger degradation than Nemotron. Without hard negatives, the L2 and L3 tiers retain $88.5\%$ and $76.2\%$ of full-resolution performance on ViDoRe v3, compared to $92.7\%$ and $91.5\%$ for Nemotron. We attribute this gap to ColPali's lower-dimensional embeddings ($128$ dimensions versus $2560$ for Nemotron), which provide less capacity to represent page-level information after heavy pooling.

\paragraph{3) Hard negatives improve robustness under aggressive compression.}
Training with mined hard negatives improves the compressed tiers, particularly at higher compression levels. On ViDoRe v3, hard negatives improve L2 and L3 retention from $88.5\%$ to $89.4\%$ and from $76.2\%$ to $80.7\%$, respectively. The gains are greater in ViDoRe v2, where L3 retention improves from $68.8\%$ to $83.4\%$. These results suggest that hard negatives help \sys learn more discriminative compressed representations when the available representation capacity is limited.

These observations address \textbf{RQ2}: \sys transfers to a smaller, lower-dimensional backbone, maintaining lossless performance up to 7.6× compression despite the limited capacity of its 128-dimensional embeddings.

\begin{table}[t]
  \centering
  {\small
  \setlength{\tabcolsep}{2.5pt}
  \begin{tabular}{@{}ll ccc ccc@{}}
    \toprule
    & & \multicolumn{3}{c}{nDCG (\%)} & \multicolumn{3}{c}{Retention (\%)} \\
    \cmidrule(lr){3-5}\cmidrule(lr){6-8}
    Model & Level & v1@5 & v2@5 & v3@10 & v1 & v2 & v3 \\
    \midrule
    ColPali v1.3 & L0 & 84.0 & 54.5 & 44.6 & 100.0 & 100.0 & 100.0 \\
    \midrule
    \multirow{4}{1.5cm}{ColSNAP-ColPali v1.3}
      & L0 & 86.1 & 54.8 & 46.9 & 102.5 & 100.6 & 105.2 \\
      & L1 & 84.1 & 53.5 & 44.5 & 100.2 &  98.0 &  99.8 \\
      & L2 & 80.6 & 46.4 & 39.5 &  96.0 &  85.0 &  88.5 \\
      & L3 & 74.7 & 37.5 & 34.0 &  89.0 &  68.8 &  76.2 \\
    \midrule
    \multirow{4}{1.5cm}{ColSNAP-ColPali v1.3-hard negs}
      & L0 & 86.4 & 54.7 & 47.1 & 102.8 & 100.3 & 105.6 \\
      & L1 & 84.7 & 52.8 & 44.7 & 100.9 &  96.9 & 100.1 \\
      & L2 & 81.9 & 49.1 & 39.9 &  97.5 &  90.1 &  89.4 \\
      & L3 & 77.3 & 45.4 & 36.0 &  92.0 &  83.4 &  80.7 \\
    \bottomrule
  \end{tabular}}
  \caption{Performance of \sys on ColPali v1.3 backbone on ViDoRe Benchmarks (V1/V2/V3). Per-tier retrieval quality and
  retention (\%), relative to the untrained baseline's L0. ViDoRe v1 and v2 are
  reported as nDCG@5, and ViDoRe v3 as nDCG@10. The top row is the untrained ColPali
  v1.3 baseline's full-resolution (L0) score.}
  \label{tab:colpali_vidore}
\end{table}

\subsection{\sys Achieves Rapid Convergence}
\begin{figure}[t]
  \centering
  \includegraphics[width=0.92\linewidth]{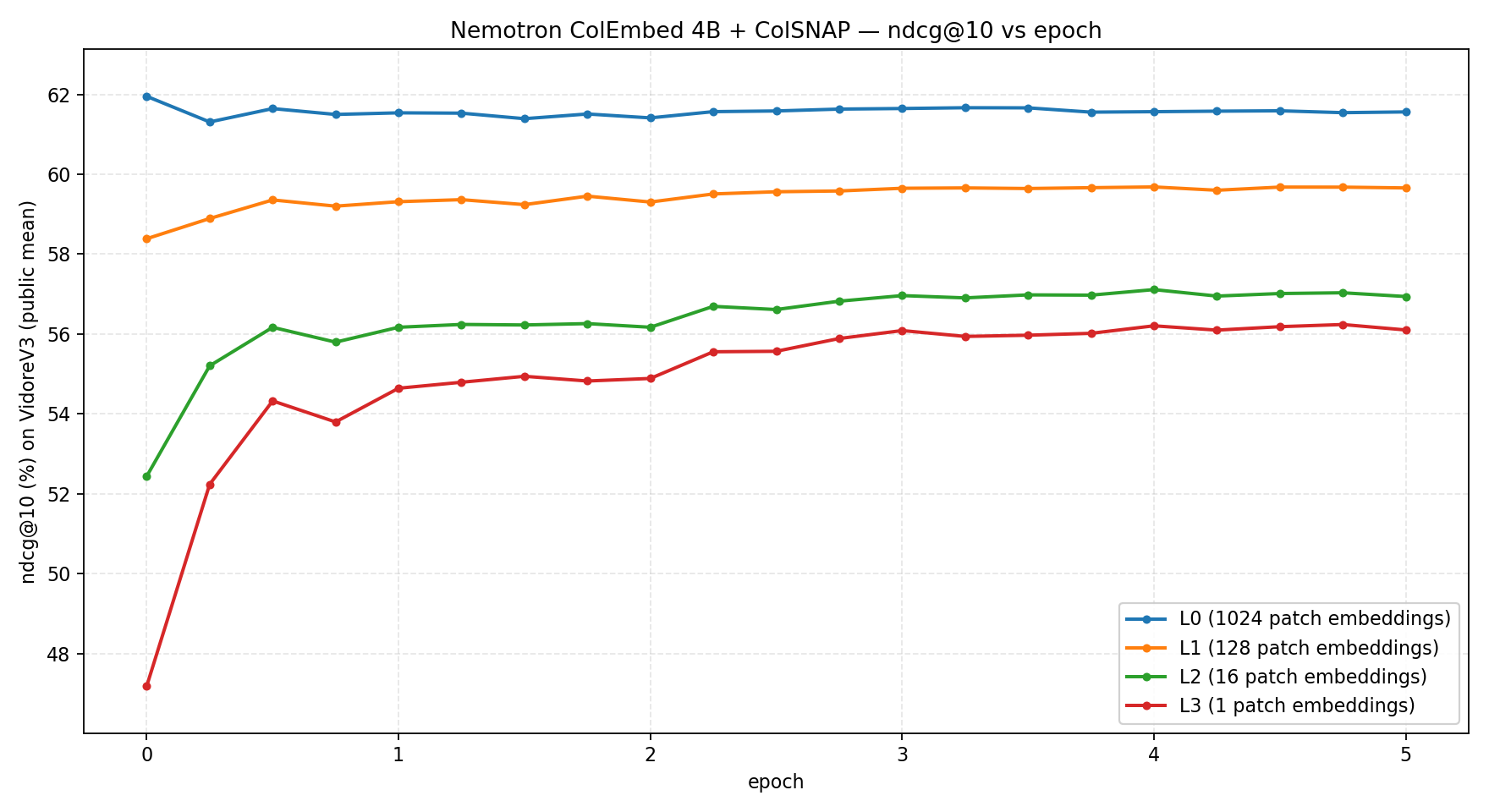}\\[4pt]
  \includegraphics[width=0.92\linewidth]{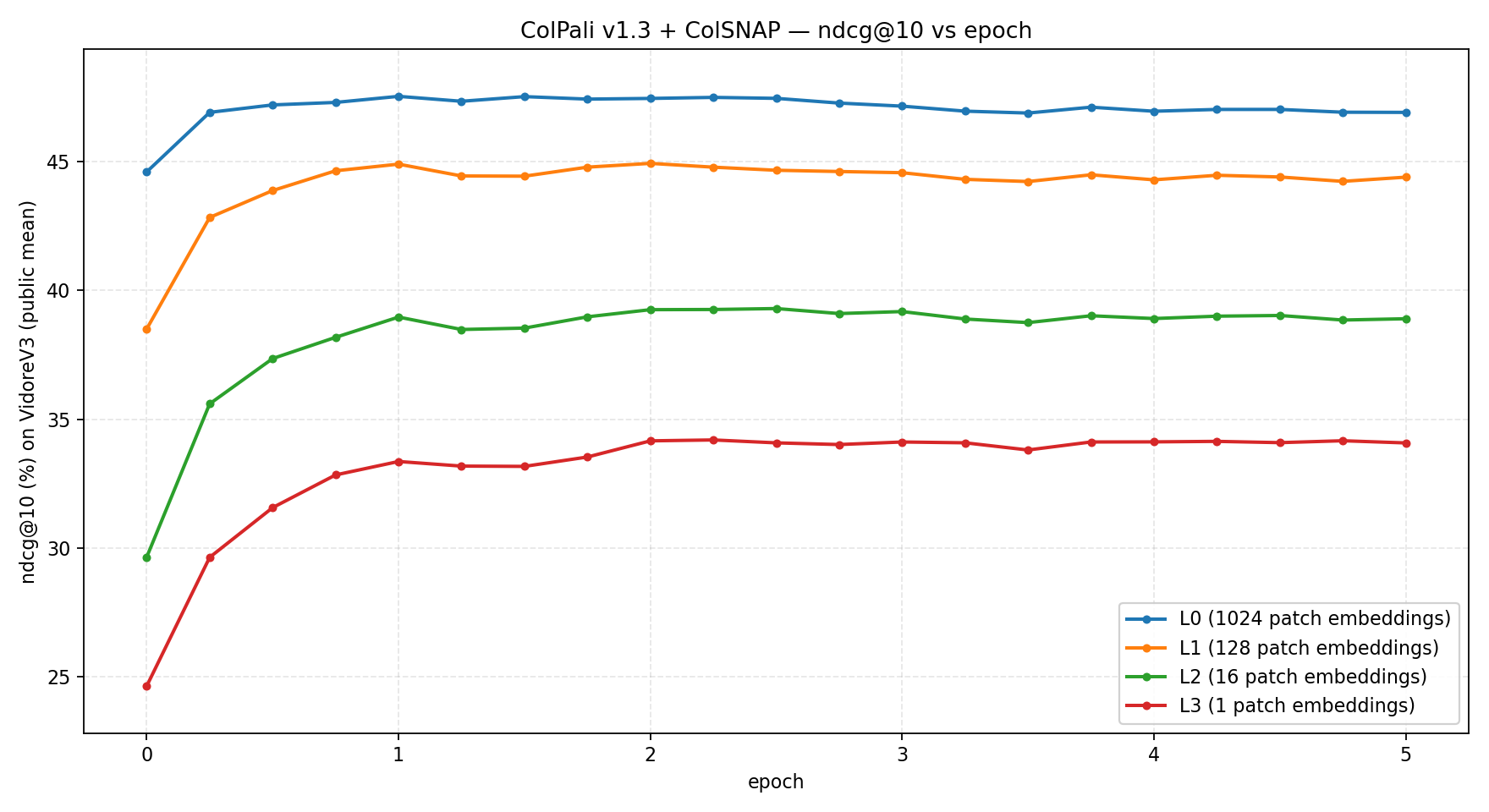}
  \caption{Per-tier nDCG@10 on ViDoRe V3 vs.\ epoch for both Nemotron ColEmbed 4B
  (a) and ColPali v1.3 (b). Step $0$ is the untrained baseline.}
  \label{fig:checkpoints}
\end{figure}

Figure~\ref{fig:checkpoints} highlights the efficiency of \sys as a post-training adaptation method. We plot per-tier nDCG@10 on ViDoRe v3 across training epochs for both backbones. The curves show that \sys converges quickly across all four tiers, with most gains achieved within the first 1--3 epochs, indicating that the model rapidly learns effective representations for coarse-grained retrieval.

On applying \sys on Nemotron ColEmbed 4B backbone, most of the improvement at compressed tiers is obtained within the first half epoch where the L1 tier reaches $59.36$, within $0.3$ nDCG points of its final score ($59.66$), while L2 and L3 achieve $83\%$ and $80\%$ of their total improvement over the untrained baseline, respectively. Further training mainly improves the coarser tiers, reaching final (epoch 5) scores of $56.94$ and $56.10$, while L0 remains stable throughout (varying by less than $0.5$ nDCG points).

\sys on ColPali v1.3 backbone follows a similar trend, converging even faster at finer tiers. After one epoch, L1 and L2 already match their final five-epoch performance, while L3 reaches $33.36$, within $0.8$ nDCG points of its final score ($34.08$). These results show that \sys can efficiently adapt existing late-interaction retrievers into multi-granularity models with only a short post-training phase.

\begin{table}[!tb]
  \centering
  {\small
  \setlength{\tabcolsep}{3pt}
  \begin{tabular}{@{}llrrr@{}}
    \toprule
    Backbone & Level & MB/1k & GFLOPs & nDCG@10 \\
    \midrule
    \multirow{4}{1.7cm}{ColSNAP-Nemotron ColEmbed 4B-hard negs}
      & L0 & 5314.56 & 187.07 & 61.6 \\
      & L1 &  727.04 &  25.59 & 60.0 \\
      & L2 &  153.60 &   5.41 & 57.7 \\
      & L3 &   76.80 &   2.70 & 57.0 \\
    \midrule
    \multirow{4}{1.7cm}{ColSNAP-ColPali v1.3-hard negs}
      & L0 & 263.94 & 8.89 & 47.1 \\
      & L1 &  34.56 & 1.16 & 44.7 \\
      & L2 &   5.89 & 0.20 & 39.9 \\
      & L3 &   2.05 & 0.07 & 36.0 \\
    \bottomrule
  \end{tabular}}
  \caption{Index size (MB per $1{,}000$ pages) and query-time scoring cost (GFLOPs
  per query) per tier, for each backbone trained with ColSNAP with mined hard
  negatives, alongside retrieval quality (nDCG@10, \%) on ViDoRe v3. Documents are
  embedded in bf16 precision; GFLOPs assume a mean query length of $35.2$ tokens for
  Nemotron and $33.7$ for ColPali (ViDoRe v3 FinanceEn).}
  \label{tab:efficiency}
\end{table}

\subsection{\sys Reduces Storage and Retrieval Cost}

Table~\ref{tab:efficiency} summarizes the storage, scoring cost, and retrieval quality trade-offs for each backbone and tier on a 1{,}000-page corpus from ViDoRe V3 with 309 queries from the \emph{finance\_en} split of ViDoRe V3. On Nemotron, the L1 tier provides the best balance, reducing index size and per-query scoring cost by $7.3\times$ while retaining $96.9\%$ of full-resolution retrieval quality. The coarser tiers deliver further reductions in storage and scoring cost, with the most aggressive setting still retaining over $92\%$ of full-resolution quality.

ColPali exhibits a similar trend: the finer tiers achieve substantial efficiency gains with negligible impact on retrieval quality, with both L0 and L1 exceeding the untrained baseline. Quality degradation emerges only at the most aggressive compression levels, reflecting the limited capacity of highly compressed representations. Overall, \sys enables significant reductions in retrieval overhead while preserving strong retrieval performance across backbones.

\section{Conclusion}
\label{sec:conclusion}
We presented ColSNAP, a training recipe that turns a pretrained late-interaction
retrieval model's own patch grid into a nested hierarchy of retrieval granularities,
without adding tokens or changing the encoder architecture. It gives a single retriever
a range of vector cardinalities to choose from at deployment time, letting users trade
compression for accuracy on demand rather than committing to one fixed
representation.

On a strong retrieval model, Nemotron ColEmbed 4B, we retained $96.9\%$ of
full-resolution retrieval quality at a $7.3\times$ reduction in stored embeddings,
and still held $93.1\%$ and $92.0\%$ at the more extreme $34.6\times$ and
$69.2\times$ reductions. On a smaller ColPali v1.3 model, the same recipe retained
lossless performance at a $7.6\times$ compression, while holding $89.4\%$ and
$80.7\%$ at $44.8\times$ and $128.9\times$. In both cases, most
of this benefit is available from a short, targeted training run, making ColSNAP a
practical option as a lightweight adaptation step, one that, with minimal additional
training, turns an existing late-interaction retriever into a multi-granularity one
whose compression level is chosen at index time rather than fixed in advance.

\bibliography{references}

\end{document}